\documentclass[12pt,twocolumn]{article}
\usepackage{amsfonts}
\usepackage{amsmath}
\usepackage{graphicx}
\usepackage{subfigure}

\title{\vspace*{-.3in}\Large \bf Comparison of Binary Classification Based on Signed Distance Functions with Support Vector Machines}

\author{\bf Erik M. Boczko\footnote{
Department of Biomedical Informatics,
Vanderbilt University Medical Center,
Nashville, TN 37232,
erik.m.boczko@vanderbilt.edu}
\and
\bf Todd R. Young\thanks{Corresponding author.
Dept.~Math,
Ohio Univ.,
Athens, OH 45701,
young@math.ohiou.edu}
\and
\bf Minhui Xie\footnote{Department of Biomedical Informatics,
Vanderbilt University Medical Center,
Nashville, TN 37232,
minhui.xie@vanderbilt.edu}
\and
\bf Di Wu\footnote{
Department of Electrical Engineering \& Computer Science,
Vanderbilt University,
Nashville, TN 37235,
di.wu@vanderbilt.edu
}}

\begin{document}

\date{}

\maketitle

\thispagestyle{empty}



\bigskip
\noindent
{\large \bf 1 \ Introduction}
\medskip

\noindent
Efficient and accurate computational solutions for binary classification
problems are currently of interest in many contexts, particularly in biomedical
informatics and computational biology where
the interesting genomic and proteomic data sets are imbued with dimensional
complexity and confounded by noise. Over the past several years it has been
effectively demonstrated that binary classification of genomic and proteomic
data can be used to connect a molecular snapshot
of an individual's internal state with the presence or absence of a disease.
This potential promises to revolutionize personalized medicine and
is fueling the development and analysis of robust classification algorithms.
Among the existing classification algorithms Support Vector Machine
(SVM) methods have distinguished themselves as efficient, accurate and
robust. Applications of Radial Basis Function Networks (RBFN) to classification
have also generated attention.

We consider only a geometric (rather than
statistical) formulation of the binary classification problem.
Namely, we suppose that the space of measurements $X$ is divided
into two subsets, $A$ and its compliment $A^c = X \setminus A$. We are given
data $\{x_i\}$ for which we know the membership in $A$ or $A^c$ of each
data point. From this data the binary classification problem is
to construct a rule or {\em classifier} that we can use to predict
the class of new, uncharacterized data. As an example, the data
may be measurements of genomic activation levels, one class might
be measures from individuals known to have a certain type of
disease while the other class may be from individuals without
the disease.

The linear SVM method was originally designed to be geometric and
robust through a constraint that it produce a
dividing surface of maximal margin between data of opposite
type. However, it has been shown that nonlinear SVM implementations
actually are built around reconstruction of the indicator function that ties
the location of the data to its class ([2]). The indicator function:
$$
    i_A(x)=\left\{\begin{array}{cl}
        1 & \mbox{if $x\in A$}\\
        -1 & \mbox{if $x\in A^c$,}
    \end{array} \right.
$$
encodes only the most primitive geometric information.
In \cite{BY} we proposed an alternative tool for classification,
the Signed Distance
Function (SDF), that measures the signed distance from the data to
the boundary between the classes, i.e.
\begin{equation}
    b_A(x)=\left\{\begin{array}{cl}
        d(x,A^c) &  \mbox{if $x \in A$}\\
        -d(x,A) &   \mbox{if $x\in A^c$,}
    \end{array} \right.
\end{equation}
where $d$ is a distance function. While the SDF has not previously
been applied to classification, it has been an important
tool in other fields, such as free boundary problems in fluid dynamics,
and so has a rich mathematical development that could be exploited.
We have tested rudimentary classification algorithms based
on the idea of reconstructing the SDF from training data. We note
that this reconstruction could be based on any accepted method
of regression, including SVM or RBFN regression. Thus, new
SVM or RBFN classification methods could be built on the SDF foundation.
One simple, yet appealing, choice
for the nonlinear regression is the least squares regression
discussed in~\cite{PS}. We have implemented this approach in
the algorithm for nonlinear data describe below.

We investigate the performance of a simple SDF based
method by direct comparison with standard SVM packages, as well
as K-nearest neighbor and  RBFN methods.
We present experimental results comparing the SDF approach with
other classifiers on both synthetic geometric problems and five benchmark
clinical microarray data sets. On both geometric
problems and microarray data sets, the non-optimized
SDF based classifiers perform just as well or slightly better
than well-developed, standard SVM methods. These results demonstrate
the potential accuracy of SDF-based methods on some types of problems.


\medskip
\noindent
{\bf Algorithms.} A procedure for training an SDF classifier
is given in Algorithm 1. It takes as arguments a
training data set $\{(x_i,y_i)\}_{i=1}^N$ and a smoothing
parameter $\gamma$, and returns a trained SDF classifier
\[
    \hat{B}(x)=\mbox{sign}[\hat{b}_A(x)]=\mbox{sign}[\sum_{i=1}^N\hat{\alpha}_iK(x,x_i)].
\]
In the SDF paradigm the input training data are marked
as to class, but they do not come marked with the values $b_A(x_i)$,
and hence these need to be approximated.
A reasonable and simple first approximation of $b_A$ at
$\{x_i\}_{i=1}^m$ is given by
\begin{equation}\label{eq7}
 b_i =  i_A(x_i) \cdot \min_{j \neq i}
     \{  d(x_i,x_j): i_A(x_j) \neq i_A(x_i) \},
\end{equation}
i.e. the signed projection onto the data of opposite type.

\begin{table}[h!]
        \begin{tabular}{p{7.7cm}}
            \hline
            \hline
            \textbf{Algorithm 1}\\
            \hline
            \hline\\
            1. For each $1\leq k\leq d$, compute the Pearson's correlation coefficient $w_k$
            between $(y_1,y_2,...,y_N)^T$ and
            $(x_{1k},x_{2k},...x_{Nk})$.\\
            \\
            2. Calculate the weighted distance matrix $\mathbf{D}$, with
            $D_{ij}=\sqrt{\sum_{k=1}^d[w_k(x_{ik}-x_{jk})]^2}$ for any $1\leq i,j\leq N$.\\
            \\
            3. Estimate the variance of the Gaussian kernel function, $\sigma$, by the Root Mean Squared Distance (RMSD)
            $\hat{\sigma}=\sqrt{\frac{2}{N(N+1)}\sum_{i=1}^{N-1}\sum_{j=i+1}^ND_{ij}^2}$.\\
            \\
            4. Calculate the Gaussian kernel matrix $\mathbf{K}$, with
            $K_{ij}=K(x_i,x_j)=\exp(-D_{ij}^2/2\sigma^2)$ for any $1\leq i,j\leq N$.\\
            \\
            5. Estimate $\mathbf{b}=\{b_i\}_{i=1}^N$ at the training data
            $\{x_i\}_{i=1}^N$ using (\ref{eq7}).\\
            \\
            6. Reconstruct $b_A$, on the entire domain through regression, i.e. solving the linear system of equations
            \begin{equation}\label{eq9}
                (\mathbf{K}+N\gamma\mathbf{I})\mathbf{\alpha}=\mathbf{b}
            \end{equation}
            where $\mathbf{I}$ is the identity matrix.\\
            \\
            \hline
            \hline
        \end{tabular}
\end{table}

In the preliminary explorations on microarray data, we used
the Pearson's correlation coefficients to rank the relevant
importance of the features (Step 1), and use these to compute
the weighted distances between cases
(Step 2). The parameter $\sigma$ determines the width of the
Gaussian functions centered at the data points.
In Algorithm 1, $\sigma$ was determined based on mean data
distances (Step 3). The Gaussian kernel matrix was calculated from the
distance matrix just in the same way as in conventional SVM
algorithms (Step 4). The SDF was estimated using the simple
heuristic improvement of equation (2). The problem of
solving (\ref{eq9}) is well-posed since
$(\mathbf{K}+N\gamma\mathbf{I})$ is strictly positive definite and
the condition number will be good provided that $N\gamma$ is not
too small (Step 6).

The procedure for training the SDF classifier on linearly
separable data is simpler and can be obtained by removing Step 2
and 3 from Algorithm 1, and replacing the Gaussian kernel function
$K(x_i,x_j)$ by the inner product operator $\langle
x_i,x_j\rangle=x_i^Tx_j$.



\bigskip
\noindent
{\large \bf 2 \ Experimental Results}
\medskip

\noindent
In \cite{BY}
we investigated these methods in linearly separable problems,
the 4 by 4 checkerboard problem and two cancer diagnosis problems
involving micro-array data. In this manuscript we report
tests of primitive SDF based methods on two new geometric problems
and four additional cancer diagnosis problems.
We compare our algorithm with the
standard SVM methods, as well as the Lagrangian
SVM~\cite{Man01} and Proximal SVM~\cite{Fun01}. A software
implementation of our algorithm with GUI can be found at:
http://people.vanderbilt.edu/~minhui.xie/sdf2005/.


\medskip
\noindent
{\bf Checkerboard Problems.}
In \cite{BY} we presented numerical results on the 4 by 4 checkerboard
problem, a geometric problem that is known to
be hard. In those tests SDF based classification
outperformed the best reported SVM results as well as
standard SVM packages.

Here we perform a simple experiment that sheds light on the
performance advantage of the SDF methods in this geometrical context.
In this test we divide the
square $X:=[-1,1]\times[-1,1]$ into two sets, $A:=[0,1]\times[0,1]$ and
its complement. We choose a data set uniformly at random in $X$ and
solve equation (\ref{eq9}) setting the right hand side to:
(a) the exact distances to the boundary $b$, (b) the exact values
of the indicator function $I$. We then use the coefficients determined in each case
to compute the value of the function $|B(0,0)|$, whose value is a
direct measure of the error in fitting the boundary between the
classes at the corner. We considered 100 independent trials and
calculated the mean value of the absolute error and its variance
(Table~\ref{tab1}).  The SDF average local error was an order
of magnitude lower as was its standard deviation, and as the number of
input data increases it decreases rapidly toward zero.
This demonstrates the main advantage of the SDF over the indicator
function: the SDF is much more suitable for the regression step.
\begin{table}[h!]\small
        \begin{tabular}{|c|c|c|}
            \hline
            Data Set  &  SDF Error     &  IF Error     \\ \hline
            100 points &  $0.0365\pm.0268$ & $0.5180\pm.2530$  \\
            500 points &  $0.0135\pm.0104$ & $0.4916\pm.1066$  \\
            1000 points & $0.0098\pm.0084$ & $0.4775\pm.0996$ \\
            \hline
        \end{tabular}
    \caption{A comparison of the local error and reliability of the
    SDF and indicator function (IF) regression applied to a local
    checkerboard problem. } \label{tab1}
\end{table}

\medskip
\noindent
{\bf Biased distribution of data.}
We observed in the linear tests that the SDF classification
particularly outperformed the SVM methods on skewed data sets.
We give here an explanation that illustrates a clear advantage
of using the SDF over the indicator function.
Suppose that the data has more samples of one type than
the other. If the indicator function is approximated then
the additional data of one type will reinforce that value
of the indicator function effectively enlarging the region
predicted in that set. If the signed distance function
is approximated, the distance to the boundary is reinforced
which does not move the boundary.

We illustrate this with a simple example. Consider the
data set $ \{ (0,1), (.1,1), (-.1,1), (0,-1) \}$.
The original SVM would place the separating line for this data
at $y = 0$. However, the Proximal SVM places it at $y = -.2$
and the Lagrangian SVM places it at $y = -.125$. If more points
are added near (0,1), the separating line is pushed further
into $y<0$. However, the SDF linear classifier places the
separating line at $y = 0$ up to an error comparable to machine
epsilon.


\medskip
\noindent
{\bf Microarray Data Sets.}
In \cite{BY} we tested an SDF-based
classifier on two standard genomic data sets involving cancer
diagnosis from microarray experiments and found the SDF-based
classification to do as well or better than standard LIBSVM
routines. In the current paper we further compare the
generalization performances of SDF-based classifier versus three
other types of distance-based classifiers: KNN, Radial Basis
Function Networks (RBFN) and SVM. We use the following microarray data
sets:
\begin{itemize}
    \item The \emph{Breast Cancer} data set~\cite{Wes01} consists
    of 49 tumor samples with 7129 human genes each. There are two
    different response variables in the data set: one describes
    the status of the estrogen receptor (ER), and the other one
    describes the status of the lymph nodal (LN). Of the 49
    samples, 25 are ER+ and 24 are ER-, 25 are LN+ and 24 are LN-.
    \item The \emph{Colon Cancer} data set~\cite{Alo99} consists
    of 40 tumor and 22 normal colon tissues with 2000 genes each.
    \item The \emph{Leukemia} data set~\cite{Gol99} consists of
    72 samples with 7129 genes each. Each patient represented by
    a sample has either acute lymphoblastic leukemia (ALL) or
    acute myeloid leukemia (AML). Of the 72 samples, 47 are ALL
and 25 are AML.
\end{itemize}

We tested the four classifiers in 100 independent trials on each
of the data sets. In each trial, the data set is divided randomly
into a training set and a test set according to the ratio of 2:1.
We use Gaussian kernel functions for RBFN, SVM and SDF.  We claim that the
classifiers are comparable in this setting since they are used under
exactly the same conditions: (i) They share the same training set
and test set in each trial, (ii) SVM and SDF share the same
$\gamma=10^{-7}$, (iii) SVM and SDF used the same weighted kernel matrix
in each trial, (iv) SVM and RBFN used the same
$\sigma$, which is computed in each trial by the RMSD as described
in Algorithm~1 in Section 3.4.
\begin{figure}[h!]
        \subfigure[]{
            \label{fig3a}
            \includegraphics[width=\linewidth,height=.73\linewidth]{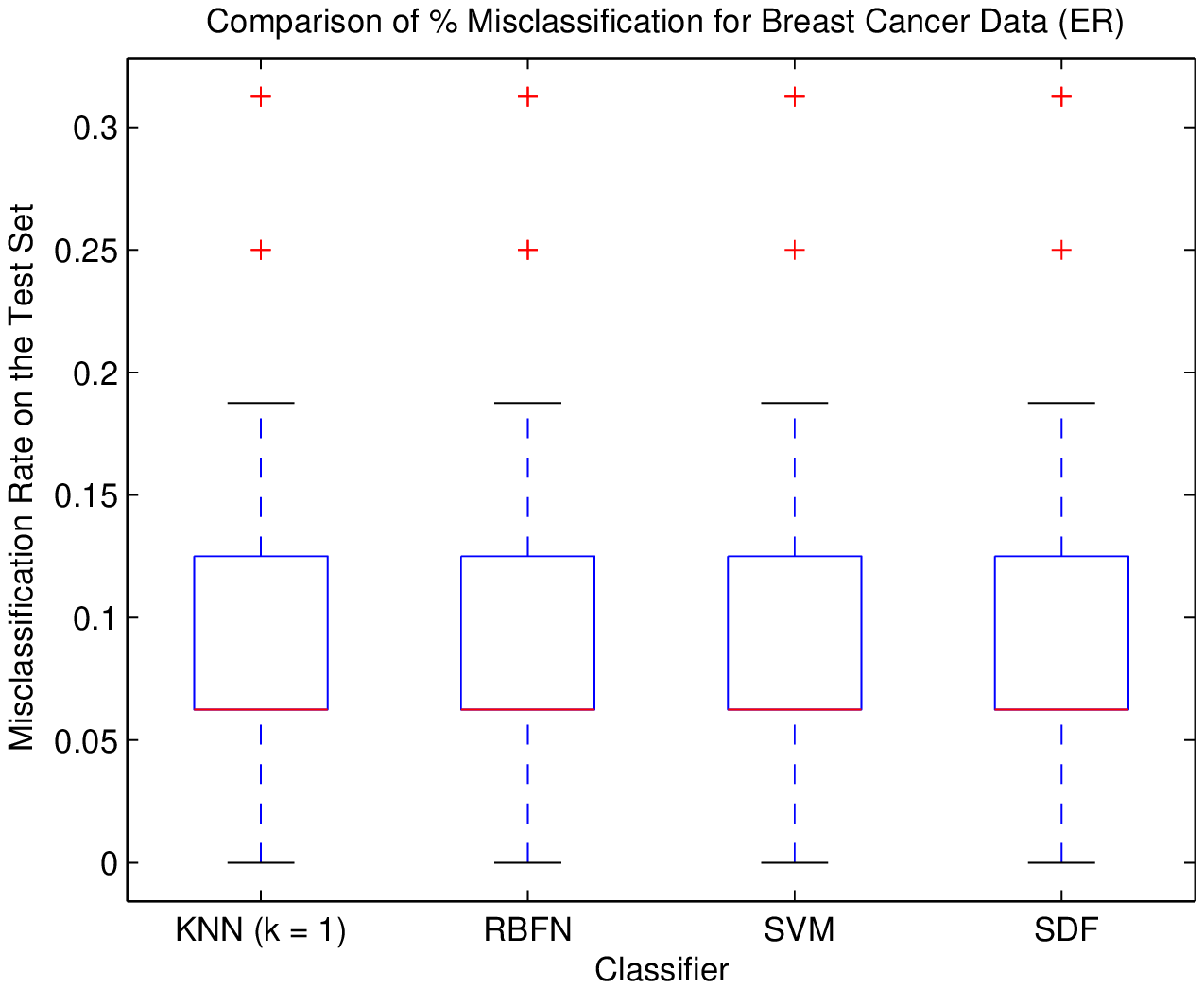}}\\
        \subfigure[]{
            \label{fig3b}
            \includegraphics[width=\linewidth,height=.73\linewidth]{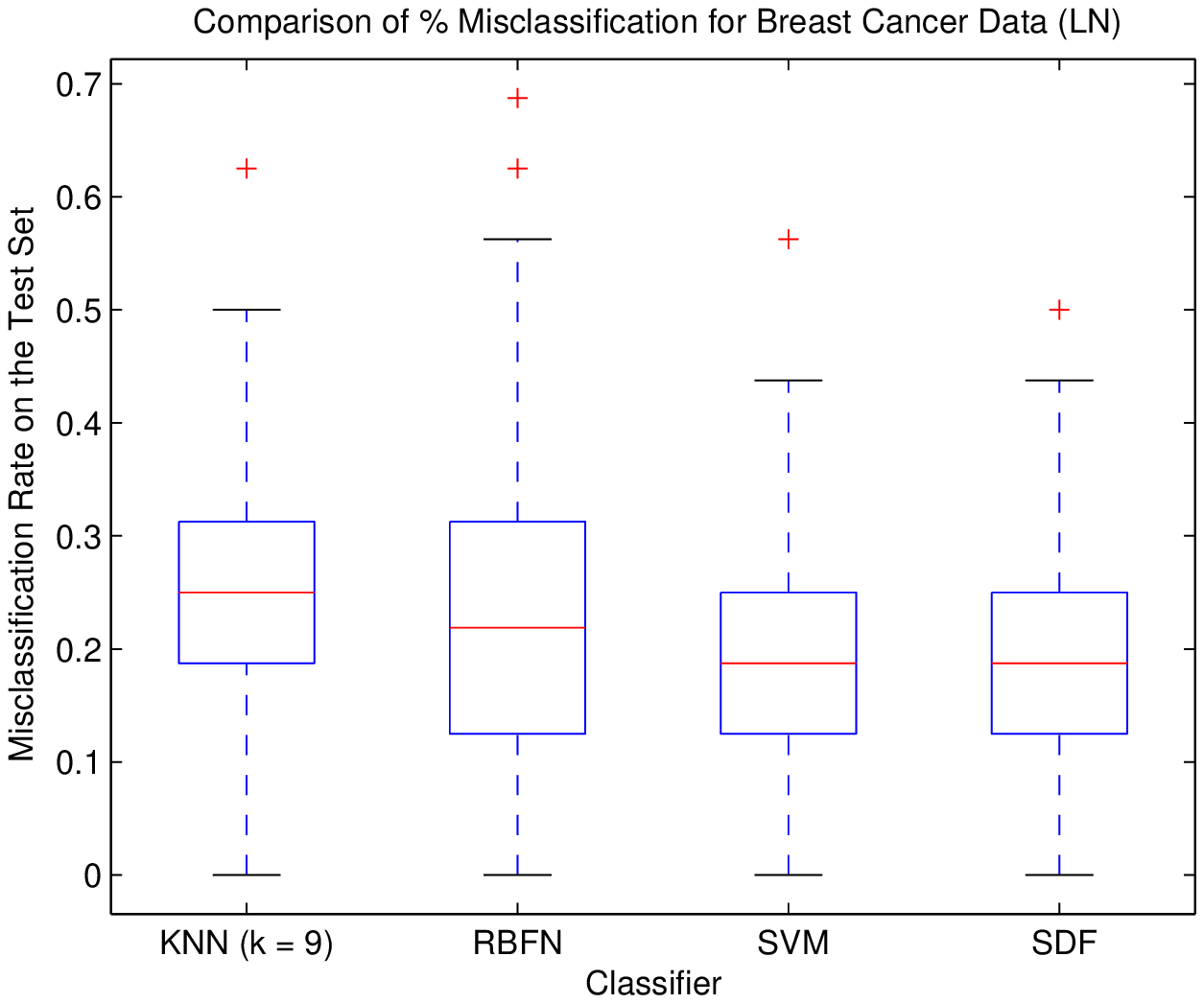}}
    \vspace*{-.2in}
    \caption{Comparison misclassification rates for breast cancer data.}
    \label{fig3}
\end{figure}

\begin{figure}[h!]
        {\includegraphics[width=\linewidth,height=.73\linewidth]{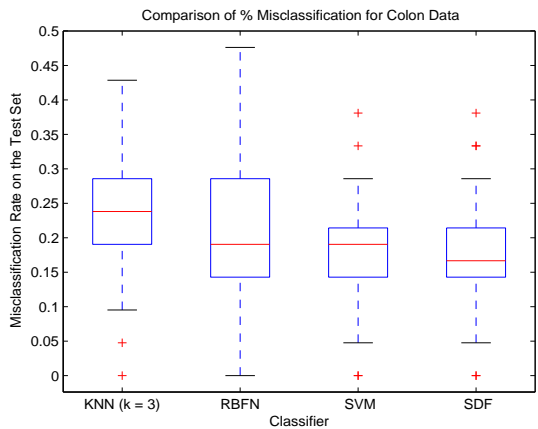}}
    \caption{Comparison of misclassification rates for colon data.}
    \label{fig4}
\end{figure}

\begin{figure}[h!]
        \includegraphics[width=\linewidth,height=.73\linewidth]{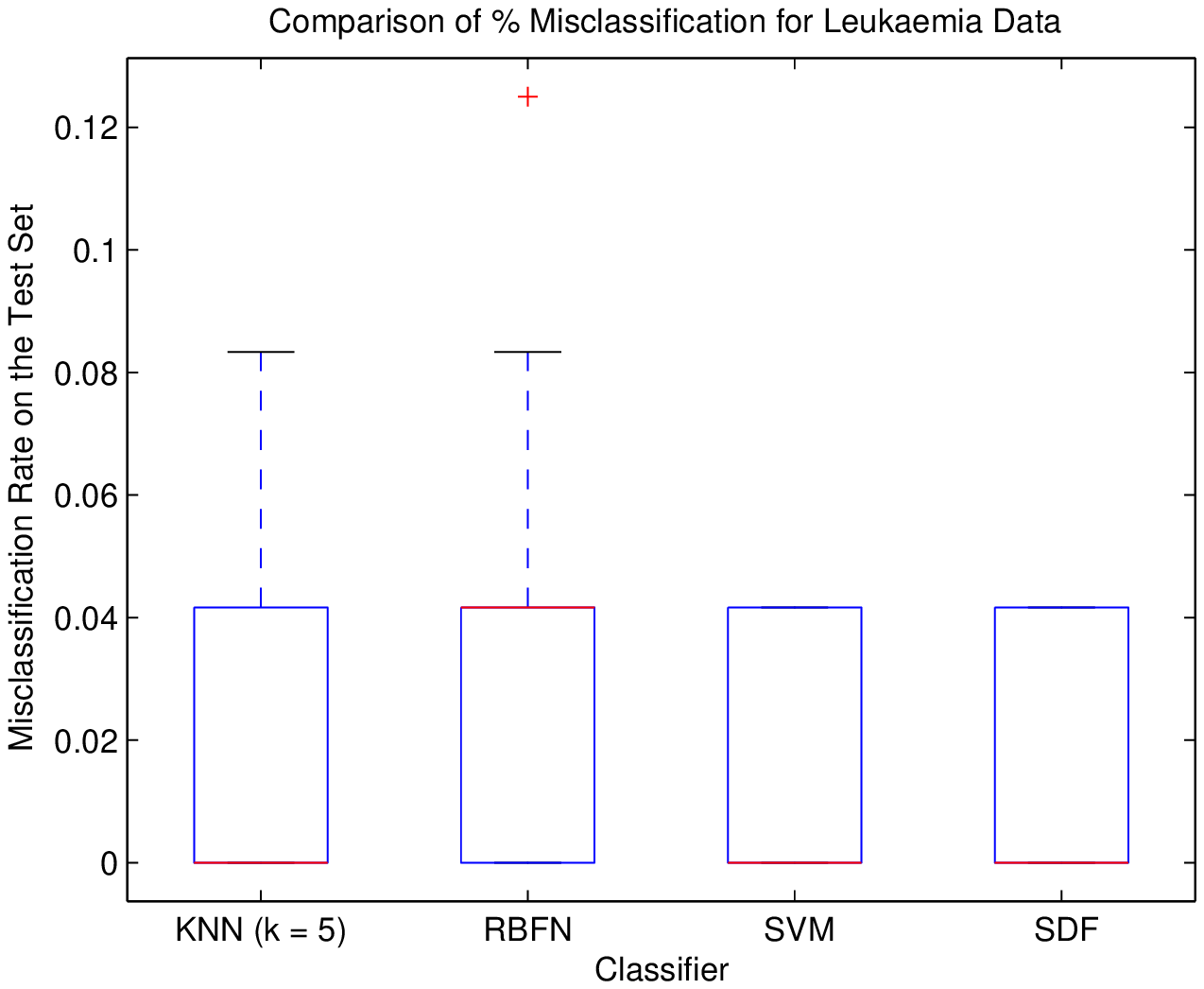}
    \caption{Comparison of misclassification rates for leukemia data.}
    \label{fig5}
\end{figure}

Figures~\ref{fig3},~\ref{fig4}, and~\ref{fig5} show the boxplots of
the test error rates in 100 trials, on the breast cancer data set,
the colon cancer data set, and the leukemia data set,
respectively. In Figure~\ref{fig3a} the response variable is ER
and in Figure~\ref{fig3b} the response variable is LN.
Since the
sample size of each data set is less than 80, KNN with a large
number of neighbors might not achieve good performance due to
overfitting. Hence in our experiments we test KNN only with $k$
from 1 to 10.

\begin{table}[h!]
    \begin{center}\small
        \begin{tabular}{|c|cccc|}
            \hline
            \multicolumn{1}{|c|}{Data Set} &\multicolumn{1}{c}{KNN} &\multicolumn{1}{c}{RBFN} &\multicolumn{1}{c}{SVM}
            &\multicolumn{1}{c|}{SDF} \\
            \hline
            Breast (ER) &.0912 &.0912 &.0869 &.0869 \\
            Breast (LN) &.2400 &.2425 &.2106 &.2100 \\
            Colon  &.2200 &.2143 &.1700 &.1662 \\
            Leukemia &.0146 &.0321 &.0167 &.0167 \\
            \hline
        \end{tabular}
    \end{center}
    \caption{Comparison of misclassification rates averaged over 100
    trials on randomly divided data.} \label{tab3}
\end{table}

Table~\ref{fig3} shows the test error rates averaged over the 100
independent trials for each classifier. KNN with 1, 9, 3, 5, 5
neighbors achieves the best (in the averaging sense)
generalization performance for the breast cancer data (ER), breast
cancer data (LN), colon cancer data, colon cancer data with 5
samples removed, and the leukemia data, respectively. In the
table we only keep the averaged test error rates for the {\em best} KNN.

From the boxplots and the averaged test error rates, we can see
that the performances of KNN and RBFN on the microarray data sets
are generally not as good as those of SVM and SDF. The performance
of the SDF classifier matches that of the SVM method on the breast
cancer data (ER) and the leukemia data, exceeds it on the breast
cancer data (LN) and the colon cancer data.



\bigskip
\noindent
{\large \bf 3 \ Conclusions}
\medskip

\noindent
From the experimental results shown above, SDF-based classifiers
promise to be more accurate than current generation
classification methods and just as efficient computationally.
Because it is geometrical, SDF-based nonlinear classification is
theoretically a more faithful and natural generalization of the
original SVM concept than existing nonlinear SVM implementations.
The algorithm presented in the paper and used in the experiments
is the most naive version, so there are many other directions of
future development that need to be followed. These might include
investigation of different methods for initialization of
$\{b_i\}_{i=1}^m$ from the training data,
optimization of $b_A$ through various iteration schemes,
exploration of the use of different regression techniques,
such as the SVM and RBFN regression, to
reconstruct $b_A$ from $\{b_i\}_{i=1}^m$, and so on. Due to the
importance of biological data sets which usually have thousands
of features, we would also try various dimension reduction
techniques, such as Principal Component Analysis (PCA),
Independent Component Analysis (ICA), and Factor Analysis to
preprocess the data and analyze the interactions between
dimensionality and performance of the SDF-based classifiers.
Finally, as pointed out in \cite{BY} the use of the SDF in
other applications and its relationship to deep mathematical results
might be exploited to both improve implementations
and provide rigorous validation of the process.




\end{document}